\documentclass[letterpaper, 10 pt, conference]{ieeeconf}  

\usepackage{amsmath}
\usepackage{amssymb}
\usepackage{adjustbox}
\usepackage{graphicx}
\usepackage[compress]{cite}
\usepackage{cancel}
\usepackage{stmaryrd}
\usepackage{pgfplots}
\usepackage{tikz}
\pgfplotsset{compat=1.18}

\usepackage{booktabs}
\usepackage{breqn}
\usepackage{algorithm}
\usepackage{pgfplots}
\usepackage{caption}
\usepackage{subcaption}
\usepackage{algpseudocode}
\usepackage{multicol}
\usepackage{caption,subcaption}
\let\labelindent\relax 
\usepackage{enumitem}
\usepackage{setspace}
\usepackage{hyperref} 

\hypersetup{
    colorlinks=true,
    linkcolor=blue,
    filecolor=magenta,      
    urlcolor=cyan,
    pdftitle={cramp},
    pdfpagemode=FullScreen,
    }
\IEEEoverridecommandlockouts                              

\title{\textbf{Zonal RL-RRT}: Integrated RL-RRT Path Planning with Collision Probability and Zone Connectivity\\}

\author{A.M. Tahmasbi$^{1}$, M. Saleh Faghfoorian$^{2}$, Aniket Bera$^{1}$%
\thanks{$^{1}$Purdue University, USA. Email: \{atahmasb, aniketbera\}@purdue.edu}%
\thanks{$^{2}$Northeastern University, USA. Email: faghfoorian.m@northeastern.edu}
}

\begin{document}
\maketitle
\begin{abstract}
Path planning in complex environments poses significant challenges, particularly in achieving time efficiency while maintaining a fair success rate and path cost. To address these issues, we introduce a novel path-planning algorithm, Zonal RL-RRT, that leverages kd-tree partitioning to segment the map into zones while addressing zone connectivity, ensuring seamless transitions between zones. By breaking down the complex environment into multiple zones and using Value Iteration as the high-level decision-maker, our algorithm achieves a 3x improvement in time efficiency compared to basic sampling methods such as RRT and RRT* in forest-like maps. Our approach outperforms heuristic-guided methods like BIT* and Informed RRT* by 1.5x in terms of runtime while maintaining robust and reliable success rates across 2D to 6D environments. Compared to learning-based methods like NeuralRRT* and MPNetSMP, as well as the heuristic RRT*J, our algorithm demonstrates, on average, 1.5x better performance in the same environments. We also evaluate the effectiveness of our approach through simulations of the UR10e arm manipulator in the MuJoCo environment. A key observation of our approach lies in its use of zone partitioning and Reinforcement Learning (RL) for adaptive high-level planning allowing the algorithm to accommodate flexible policies across diverse environments, making it a versatile tool for advanced path planning.  



\end{abstract}

\section{Introduction}
Path planning involves computing a collision-free trajectory from a start configuration to a goal configuration within an environment containing obstacles. Numerous strategies have been developed to efficiently handle complex and highly constrained environments while remaining general enough to perform effectively across diverse settings and optimization criteria, such as travel time, path length, or maximizing safety. Current path-planning algorithms are broadly classified into several categories.
Graph-based algorithms model the environment as a network, seeking the shortest path, such as Dijkstra's algorithm \cite{b45}, A*\cite{b47}, and D* \cite{b46}. Although these methods guarantee an optimal path, their application in high-dimensional spaces incurs a significant increase in memory demands and computational expense. 
\begin{figure}[htbp]
\centerline{\includegraphics[width=1.1\linewidth]{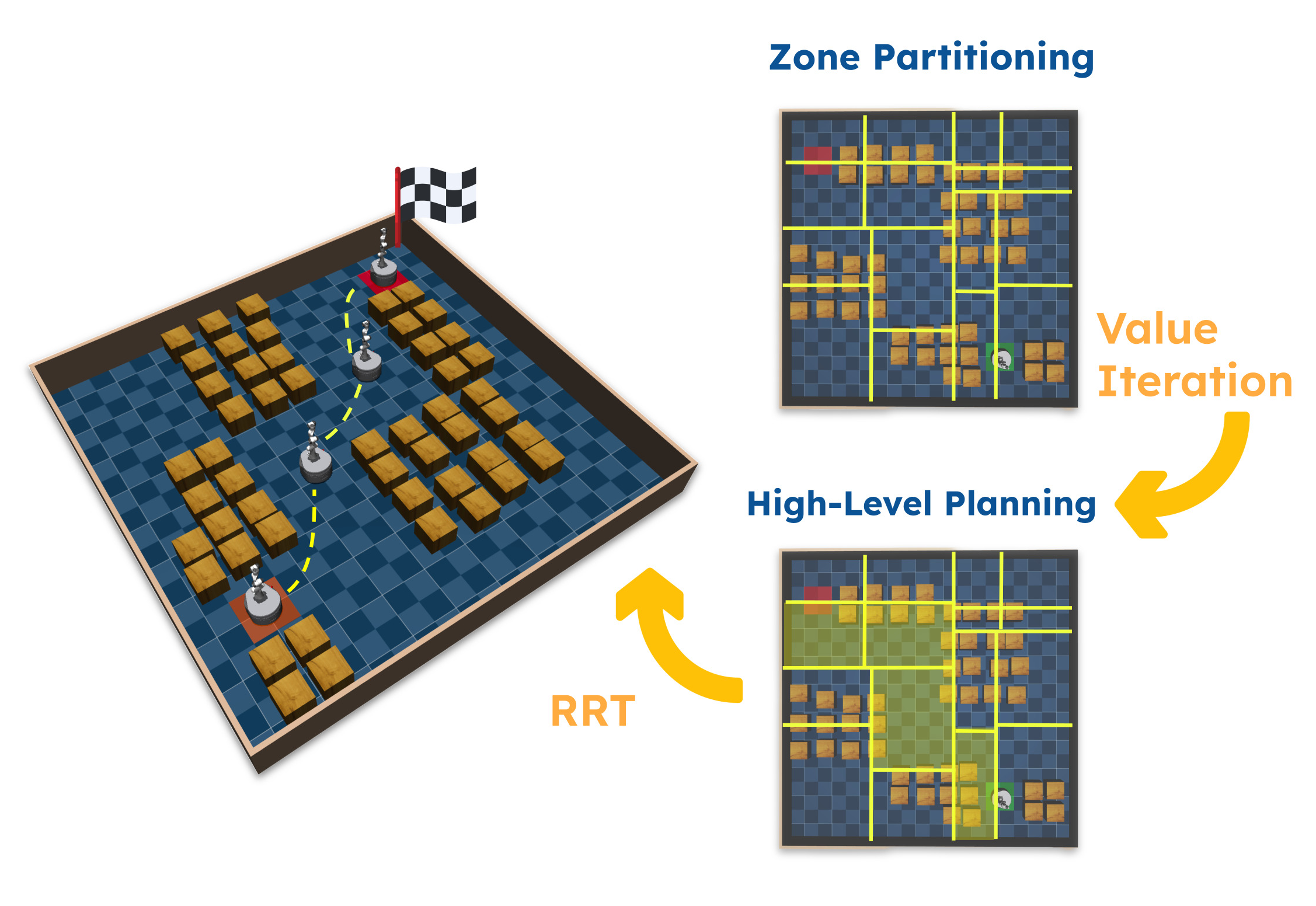}}
\caption{\small \textit{Illustration of our Zonal RL-RRT approach. The input is the map including locations of all obstacles, start location, and goal positions. We partition the map into multiple zones using a kd-tree and establishes a state-action space accounting for the connectivity between zones. Our approach determines the optimal sequence of zones to traverse from the start to the goal. Finally, we utilize the Rapidly-exploring Random Trees (RRT) method low-level path planning within these zones and generate the final route.}}
\label{coverfig}
\vspace{-0.6cm}
\end{figure}

The Artificial Potential Field (APF) method \cite{b48} generates virtual force fields to facilitate smooth navigational adjustments for agents but is significantly limited by its tendency to become trapped in local minima. Emerging learning-based solutions, including data-driven approaches that computationally enhance traditional methodologies \cite{b51,b52,b53}, have been introduced. However, these techniques often struggle with generalizability to unseen scenarios. Reinforcement Learning (RL) \cite{b19}, characterized by environment-driven learning, optimizes complex objectives with minimal oversight but requires extensive training data for effective convergence in intricate settings \cite{b1,b2,b3}. Sampling-based methods like RRT \cite{b10}, RRT* \cite{b54}, and their derivatives are popular for their efficacy in navigating high-dimensional, constrained spaces; nevertheless, their uniform random sampling nature introduces suboptimality in path generation, with solution quality contingent on the number of samples produced.

Despite these challenges, the computational advantages of RRT are significant, leading to the development of numerous hybrid sampling algorithms. These enhancements fall into two main categories. First, heuristically guided search augments traditional graph-based methods with efficient sampling techniques, such as BIT* \cite{b13}, Informed RRT* \cite{b57}, and Theta*-RRT \cite{b55}. However, issues with memory consumption and processing time persist in large and complex maps. Second, learning-enhanced approaches \cite{b24,b26,b49,b50}, particularly those trained on pre-generated paths from heuristic algorithms, rely heavily on the quality and specific constraints of their training data. Consequently, they offer limited flexibility at test time and provide little direct control over key path properties such as length or safety, as these methods typically lack explicit mechanisms for modulating planning strategies during execution.

In this paper, we present a two-phase planning framework that integrates high-level decision-making with low-level sampling. The high-level phase leverages Value Iteration to compute an optimal sequence of zones to traverse, based on a kd-tree decomposition of the environment that reflects obstacle concentration and spatial structure. This phase enables explicit control over planning behavior by tuning reward weights associated with zone transitions. Once the global zone path is determined, the low-level phase employs RRT within each zone to generate local paths between subgoals. The combination ensures a coherent and collision-free trajectory from start to goal(Figure 1). Finally, a Greedy Shortcut algorithm is applied in post-processing to smooth the resulting path by eliminating unnecessary waypoints, improving path quality. Our main contributions are as follows:

\begin{itemize}[left=0pt]
\item By incorporating the spatial distribution of obstacles and partitioning the map, we effectively reduce complexity, especially in highly dense and cluttered environments. In 2D forest maps, it achieves an average 96\% reduction in run time and improves the path cost by 5.5\% compared to the four baselines (RRT, RRT*, BIT* and Informed RRT*). In 3D environments, it delivers an 87\% runtime reduction with a 4.4\% cost improvement on average. In a 6D robotic environment using the UR10e manipulator, our algorithm demonstrates faster performance compared to BIT* on average.

\item Compared to prominent learning‑based methods such as MPNet SMP and Neural RRT*, and heuristic planners like RRT*J, Zonal RL‑RRT reduces runtime by about 47\% on average while keeping path costs within ± 3.3\% of these approaches, despite requiring no offline training.  Unlike data‑driven planners that depend on pre‑collected paths from specific environments, our algorithm requires no offline training and grants explicit control over planning behavior via tunable reward weights.

\item Using map partitioning and value iteration for high-level planning, Zonal RL-RRT offers flexible strategy adaptation. It can operate conservatively, navigating through broadly free zones to maximize safety, or greedily, selecting zones closer to the goal for faster routes simply by adjusting its reward parameters. This zone‑based framework therefore supports various planning objectives without retraining.

\end{itemize}
    
\section{Related works}
\subsection{Sampling-Based Planning Methods}
Sampling-based methods leverage continuous spaces and randomized characteristics for enhanced exploration capabilities. At the forefront of these methods stands the Rapidly-exploring Random Tree (RRT), a foundational algorithm from which many other algorithms in this field are derived\cite{b10}.
RRT efficiently navigates complex environments by rapidly exploring high-dimensional spaces while respecting obstacle constraints~\cite{b11}. Despite these advantages, basic sampling–based methods like RRT cannot consistently find optimal paths, since uniform sampling often overlooks narrow passages due to its non‑discriminative distribution~\cite{b12}. To address the inherent limitations of the basic RRT algorithm recent developments have emerged, categorizable into distinct groups.
\subsubsection{Heuristically Guided Search} Heuristically guided searches such as Informed RRT*, and Batch Informed Trees (BIT*) have enhanced the efficacy of path planning in complex environments. Theta*-RRT hierarchically merges any-angle search with RRT, guiding tree expansion toward strategically sampled states that encapsulate geometric path information\cite{b14}. A*-RRT and A*-RRT* refine this concept by sampling along A* paths, with the latter distributing samples around the path to minimize costs effectively\cite{b16}. BIT* unifies the graph-based search's ordered nature with the scalability of sampling-based methods by interpreting a set of samples as an implicit random geometric graph (RGG)\cite{b15}. Additionally, RRT*J introduces a novel non-uniform sampling technique that leverages a generalized Voronoi graph (GVG) and discretizes the heuristic path to construct multiple potential functions (MPF), which further guide exploration toward optimal solutions\cite{b60}. 
Although these approaches improve planning by directing exploration towards areas identified as promising through predefined heuristics, this focused approach can unintentionally restrict the exploration breadth, particularly in settings with intricate constraints or clutter, where the heuristic might not accurately reflect spatial complexities. Moreover, the efficiency of these methods tends to diminish with increasing map sizes or state space dimensionality, especially when they rely on detailed environmental information or precomputed data structures, which can be computationally expensive.
\subsubsection{Learning-Enhanced Sampling}
Enhanced Sampling through Learning incorporates methods that use Deep Learning, Reinforcement Learning, or Neural Networks \cite{b26,b27} to refine the sampling process, thereby increasing the efficiency of path-planning algorithms.
In \cite{b22}, they have developed a method that employs reinforcement learning techniques to enhance the sampling strategy within a discretized workspace environment. In \cite{b23}, a technique is introduced that utilizes conditional variational autoencoders (CVAEs) to create a specific sampling distribution optimized for motion planning applications. In \cite{b24}, the authors introduced MPNet, which leverages Neural Networks to derive near-optimal path planning heuristics from environmental data and robot configurations, enabling bidirectional path generation. 
In \cite{b26}, NRRT* has been proposed, utilizing a CNN model to predict the probability distribution of the optimal path. This model, trained on data generated by the A* algorithm, guides the sampling process to enhance path planning efficiency. While these learning-enhanced sampling methodologies exhibit improved performance compared to classical approaches, they are confronted with significant challenges. 
Learning-based methods trained on heuristic-guided algorithms (such as A*) often inherit the limitations of these heuristics, impeding generalization to new, unseen environments. This reliance reduces their flexibility to adapt dynamically when confronted with novel obstacles or complex environmental features.
\subsection{Partitioning Strategies}

Several partitioning strategies have been developed to enhance the zonal division of maps for path-planning purposes. Uniform Grid (UG) partitioning~\cite{b8} divides the environment into equally sized grid cells, offering straightforward implementation, but it can be memory-intensive and inefficient in sparse environments due to equal allocation of space to both occupied and unoccupied regions. In contrast, shape‑aware methods~\cite{b7,b8} adapt their zones to obstacle layout; one lightweight example is the kd‑tree~\cite{bruce2005}, which uses recursive binary splits to carve out tighter, data‑driven regions. By aligning zones with actual obstacle distributions.

\subsection{Smoothing Algorithms}
Path smoothing is widely used to improve path quality by reducing unnecessary waypoints and sharp turns. Common techniques include geometric methods such as Greedy Shortcut and Douglas-Peucker \cite{b65}, as well as curve-based approaches like Bézier and B-spline smoothing \cite{b66}. These methods vary in smoothness, computational cost, and applicability to different planning frameworks.

\section{Our Approach}
In this section, we will give an overview of our methodology, followed by technical details.
\subsection{Formulations}
\vspace{-0.1cm}
The path planning process begins with a predefined \textit{\textbf{A}}gent's start point, goal point, and the positions of obstacles within the environment space \textit{\textbf{E}}.
\vspace{-0.2cm}
\begin{equation}
A_{start} = (x_{start}^1, \dots, x_{start}^n),\quad \\ A_{goal} = (x_{goal}^1, \dots, x_{goal}^n)
\end{equation}

where \(n\) represents the number of dimensions in the environment. In this work, we assume that all obstacles are axis-aligned rectangles (in 2D) or boxes (in 3D). Each obstacle is defined by its minimum and maximum bounds along each axis.
Let the environment \textit{\textbf{E}} be partitioned into $k$ unique zones, with each zone distinguished by specific constraints and attributes. The set of all zones is denoted by $\mathbf{Z}$, where each $Z_i$ is an axis-aligned hyperrectangular region:
\begin{equation}
\small
E = \bigcup_{i=1}^{k} Z_i \quad\text{with}\quad Z_i \cap Z_j = \varnothing, \quad \forall i \neq j, \; i, j \in \{1,\ldots, k\}
\end{equation}

\subsection{Overview}
The goal is to define an optimal sequence of zones \( Z_o = \{ Z_{\text{start}}, Z_{i1}, \dots, Z_{im}, Z_{\text{goal}} \} \), where \( A_{\text{start}} \) is in \( Z_{\text{start}} \) and \( A_{\text{goal}} \) is in \( Z_{\text{goal}} \). The intermediate zones are determined using a Value Iteration framework, and RRT is subsequently employed as a local planner to generate the path within these zones.
\vspace{-0.1cm}
\subsection{Methodology}
\vspace{-0.2cm}
\subsubsection{Reward Function}
The rewards for zones are influenced by various parameters reflecting their characteristics, which depend on the specifics of the scenario. We suggest a set of fundamental criteria that should be considered in all scenarios.\cite{b1,b2,b3,b4}
\paragraph{Collision Probability ($R_{\rho}$)}
The parameter $R_{\rho}$, captures the obstacle density within a zone, indicating collision likelihood, higher values suggest increased risk, defined as:
\begin{equation}
R_{\text{$\rho$}} \propto -  A_{\text{obstacles}} / A_{\text{zone}}
\end{equation}
This metric effectively gauges the probability of collision, with denser areas posing a higher risk to the agent.
\paragraph{Distance to goal$(R_d)$}This reward component, which accounts for distance, is structured to reflect the proximity of a zone's center to the target. This encourages strategies that favor shorter, more direct trajectories across fewer zones.
\begin{equation}
R_{\text{$d$}} \propto -  \lVert Z_{center}- A_{goal} \lVert
\end{equation}
Now, by considering all parameters, the final reward function is enhanced to:
\begin{equation}
 R = \omega_1 \cdot R_{\text{$d$}} + \omega_2 \cdot R_{\text{$\rho$}} +  \omega_3  \times I[\text{goal\_reached}] 
\end{equation}
where \( \omega_1,\omega_2 ,\omega_3 \) are weighting factors for the respective components of the reward function, and the indicator function \( I[\text{goal\_reached}] \) takes a value of 1 when the goal is reached, contributing positively to the reward, and 0 otherwise.

\subsubsection{kd-tree Algorithm Formulation}
The kd-tree algorithm partitions the environment workspace \textit{\textbf{E}} into rectangular axis-aligned zones, based on the distribution and density of obstacles. This partitioning is mathematically represented as:
\begin{multline}
\small
\text{Partition}(E, \text{depth}) = \\
\begin{cases} 
   \text{Split}(E, \text{median}(C_{\text{axis}}), \text{axis}) & \text{if depth} \leq \text{MaxDepth} \\
   E & \text{otherwise}
\end{cases} 
\end{multline}

where \(\text{Split}(E, \text{median}(C_{\text{axis}}), \text{axis})\) divides \(E\) at the median of obstacle centers along the current axis. \(C_{\text{axis}}\) represents the coordinates of obstacle centers on the chosen axis, which alternates between \(x\) and \(y\) in 2D, and between \(x\), \(y\), and \(z\) in 3D, with increasing depth. The kd-tree's dynamic partitioning approach allows for adaptive zone sizing, which can be directly incorporated into the reward function by \(R_{\rho}\), as it can more accurately reflect the density of obstacles within dynamically determined zones. This leads to a reward system that is sensitive to the spatial distribution and density of obstacles, enhancing the path-planning process.

\subsubsection{Zone Connectivity}
After dividing the map into zones with a kd-tree, we establish an action state space for Value Iteration to derive an optimal high-level strategy. Zones \(i\) and \(j\) are deemed inaccessible to each other if they lack a common border or if their border is blocked by obstacles.
To verify obstacle-induced blockage, a specific procedure is employed (Algorithm 2): We delineate a narrow area along the zones' shared boundary, defined by \(\delta\), and assess the largest gap between obstacles within this region. If the maximum distance between any two adjacent points \(x_i\) and \(x_{i+1}\) falls below a set threshold $\gamma_{C}$, indicative of no viable passage, those zones are considered disconnected. This threshold should at least match the agent's radius, ensuring navigability. The order of complexity of this algorithm is \(O(n \times \log(n))\), where \(n\) represents the number of obstacle points falling within the border region between zones, which is a fraction of the total points in the map.
\subsubsection{Value Iteration}
Given the discrete maps with a limited number of zones from partitioning with fully known constraints and transitions, Value Iteration is an ideal choice for its proficiency in handling environments with well-defined, finite state spaces, thus facilitating efficient policy optimization. It updates values using the Bellman equation:

\begin{multline}
\small
V(Z_t) \gets \\
\max_{A_t} \left[ R(Z_t, A_t) + \gamma \sum_{Z_{t+1}} P(Z_{t+1} | Z_t, A_t) V(Z_{t+1}) \right]
\end{multline}

where $V(Z_t)$ represents the value of zone $Z_t$, $R(Z_t, A_t)$ is the immediate reward received by taking action $A_t$ in zone $Z_t$, and $P(Z_{t+1} | Z_t, A_t)$ denotes the transition probability to zone $Z_{t+1}$ given action $A_t$. The discount factor $\gamma$ determines the weight of future rewards. The optimal policy $\pi^*$ is derived by selecting actions that maximize the expected value function.
Algorithm 1 combines kd-tree partitioning and policy to guide an agent from start to goal. It determines zones through policy, then identifies a secure subgoal within the upcoming zone by evaluating $m$ number of potential candidates. This process involves calculating each potential next node's proximity to the obstacles within the zone and deeming a subgoal as safe if it maintains a distance greater than the safety threshold, $\gamma_s$. The order of complexity of this algorithm is $O(n)$, where $n$ represents the number of obstacles evaluated within the zone. Once the agent enters the goal zone and reaches the goal, a shortcut smoothing technique is applied to the entire path from start to goal. This technique iteratively replaces intermediate segments with direct connections, if they are collision-free, to improve the overall path quality in terms of length. The time complexity is $O(mn^2)$ where $m$ represents the number of rectangular obstacles and $n$ represents the number of waypoints in the path.
\begin{algorithm}
\small
\caption{Zone Connectivity}
\begin{algorithmic}[1]
\Procedure{IsAccessible}{$z_i,z_j,\mathcal O$} 
\If{zones have vertical common border}
    \State \textbf{Let} $P_v =$
    \State \resizebox{0.87\linewidth}{!}{ $\{(x, y) \mid |x - a| < \delta, y \in $ common $y$ range$, (x,y) \in \text{ $\mathcal O$}\}$}
    \State \textbf{sort} $P_v$ based on y-coordinate
    \If{$\max_{(p_i, p_{i+1}) \in P_v} \lVert p_i - p_{i+1} \rVert \quad < \gamma_{C} $}
    \State \textbf{return} False
    \EndIf
\EndIf
\If{zones have horizontal common border}
\State \textbf{Let} $P_h =$
    \State =\resizebox{0.87\linewidth}{!}{ $\{(x, y) \mid |y - a| < \delta, x \in $ common $x$ range$, (x,y) \in \text{ $\mathcal O$}\}$}
    \State \textbf{sort} $P_h$ based on x-coordinate
    \If{$\max_{(p_i, p_{i+1}) \in P_h} \lVert p_i - p_{i+1} \rVert \quad < \gamma_{C} $}
    \State \textbf{return} False
    \EndIf
\EndIf
\State \textbf{return} True
\EndProcedure
\end{algorithmic}
\vspace{-.1cm}
\end{algorithm}
\vspace{-18pt} 
\begin{algorithm}
\small
\caption{Get Trajectory}
\begin{algorithmic}[1]
\Procedure{Path}{$\mathcal O,\mathcal Z, policy \ \pi,\text{goal},\text{start}$} 
\If{$NextZone = GoalZone$}
    \State $path \leftarrow RRT\*(start, goal, Obstacles)$
    \State $path \leftarrow ShortcutSmoothing(path, \mathcal{O})$
    \State break
\EndIf
\While{$NextZone \neq GoalZone$}
    \State $subgoal \gets \Call{Choose SubGoal}{nextGrid}$
    \State $ path \gets RRT(start,subgoal,\mathcal O)$
    \State $NextZone \gets Policy \ \pi$
    \State $start \gets subgoal$
\EndWhile
\EndProcedure
\end{algorithmic}
\end{algorithm}
\section{Evaluation}
This section presents a comprehensive evaluation of our proposed algorithm, demonstrating its performance across diverse environments and algorithmic configurations, such as varying kd-tree depths, and providing comparisons against established path-planning algorithms in 2D, 3D, and eventually deploying it on a 6DOF robot manipulator. The Value Iteration hyperparameters are set as follows: \( \gamma = 0.95 \), \( \epsilon = 1\times10^{-6} \).
Also, we utilized standard OMPL \cite{b59} implementations for classic, well-known planners, including RRT, RRT*, BIT*, and Informed RRT*.

\begin{table}[h]
\centering
\caption{Performance Comparison of Path Planning Algorithms in 2D Forest Map}
\label{my-label}
\small
\setlength{\tabcolsep}{3pt} 
\renewcommand{\arraystretch}{1} 
\begin{tabular}{@{}cclccc@{}} 
\toprule
No. Obs & Obs Size & \multicolumn{1}{c}{Algorithm} & Time(s) & Sr(\%) & cost(\%) \\ \midrule
50     & 20   & \raggedleft RRT                        & 0.15  & 100 & +16   
  \\  
       &      & \raggedleft RRT*                       & 15.08  & 100 & +12  \\  
       &      & \raggedleft BIT*                        & 0.20  & 100 & -11 
       \\  
       &      & \raggedleft Informed RRT*              & 0.55  & 100 & -11  \\  
       &      & \raggedleft \textbf{ZRL-RRT (16 zones)} & \textbf{0.06}  & \textbf{100} &   \\  
       \midrule
250     & 10   & \raggedleft RRT                        & 1.27  & 100 & +20 \\  
       &      & \raggedleft RRT*                       & 15.09  & 100 & +18  \\  
       &      & \raggedleft BIT*                        & 1.01  & 100 &  +5 \\  
       &      & \raggedleft Informed RRT*              & 6.05  & 95 & +2  \\  
       &      & \raggedleft \textbf{ZRL-RRT (16 zones)} & \textbf{0.29}  & \textbf{100} &  - \\  
       \midrule
500     & 5   & \raggedleft RRT                        & 1.01  & 87 & +9  \\  
       &      & \raggedleft RRT*                       & 40.05  & 100 & +6  \\  
       &      & \raggedleft BIT*                        & 2.52  & 100 & -12  \\  
       &      & \raggedleft Informed RRT*              & 40.05  & 100 & -18  \\  
       &      & \raggedleft \textbf{ZRL-RRT (16 zones)} & \textbf{0.76}  & \textbf{100} &  - \\  
       \midrule
\end{tabular}
\vspace{-0.5cm}
\end{table}
\vspace{-0.1cm}
\subsection{2D Environment}

First, classical sampling-based methods and Zonal RL-RRT are evaluated across a diverse set of \(200 \times 200\) 2D forest maps, each featuring varying numbers and sizes of rectangular obstacles. The evaluated methods include RRT, RRT*, BIT*, and Informed RRT*, with runtime, success rate, and path length as the three main metrics. Note that all reported times for Zonal RL-RRT include the preprocessing phase, such as zone connectivity, value iteration, and smoothing process.
These maps are consistently used across all algorithms to ensure fair and unbiased comparisons. Time thresholds were set to ensure that the algorithms could find an initial path with a fair success rate and reasonable computational demands, particularly for those without specific termination criteria, such as BIT*, which begins refining the path immediately after finding an initial solution.
Experiments conducted in Python 3.10.12 were evaluated using an Intel i7 CPU with 12 GB of RAM running on Ubuntu 22.04.4. 
In Table I, we evaluate the performance of various algorithms on forest-like maps featuring obstacle counts from 50 to 500, with obstacle sizes of 5, 10, and 20 (step size 1). Each configuration was tested over 100 iterations. 
Overall, in all three scenarios, Zonal RL‑RRT matches or exceeds the success rates of the baselines and outperforms them in terms of runtime. On average, Zonal RL‑RRT with depth 4 is about 2.2× faster than RRT and 3.4× faster than BIT*, and it also outpaces both RRT* and Informed RRT* in runtime, and in terms of path length, Zonal RL‑RRT reduces cost by an average of 13.5\% across all scenarios. When compared to informed sampling–based planners, it delivers competitive performance, with path costs within 7.5\% of theirs on average. 
In our algorithm, the depth parameter, which determines the number of zones, plays a central role in guiding the path. As shown in Figure 5, increasing the zone resolution directs sampling toward more relevant regions and helps avoid costly detours. Specifically, compared to depth 2, the mean path cost decreases by 17\% at depth 3 and by 20\% at depth 4. The modest 3\% improvement from depth 3 to depth 4, along with the fact that both remain on average within 10\% of the optimal path costs produced by planners like BIT*, suggests diminishing returns beyond three levels of partitioning.
Zone resolution also impacts the runtime distribution across different phases of the algorithm. The high-level zone connectivity phase grows from 0.02\% of total runtime at depth 2 (0.4 ms) to 0.29\% at depth 4 (2.9 ms), and the value iteration phase increases from 0.01\% (0.3 ms) to 0.26\% (2.6 ms). Beyond depth 3, further partitioning adds computational overhead without a meaningful gain in path quality. In contrast, the share of runtime spent in the local planning phase drops from 75\% at depth 2 to 61\% at depth 4, indicating that more specific sampling regions improve local planner efficiency.

Overall, these results suggest that the depth parameter should be tuned according to the complexity of the environment. While increasing the number of zones can help reduce detours and accelerate local planning, it may also introduce unnecessary overhead in the high-level planning stages. Therefore, selecting a balanced depth is crucial to achieving optimal performance.
To illustrate the differences between the three partitioning approaches, Table III presents their performance in terms of success rate and path cost. All three methods- Uniform Grid (UG), QuadTree (Qd), and kd-tree, achieve a 100\% success rate in the tested scenario. In terms of path cost, kd-tree shows improvements of 10\% and 19\% over UG and Quad-tree, respectively. Notably, increasing the number of zones in the QuadTree approach exhibits a trend similar to kd-tree: using 64 zones improves path quality by 11\%, but leads to an 8 times increase in high-level planning time (up to 18 ms, including zone connectivity and value iteration). However, this overhead remains below 1\% of the total planning time. These results suggest that different zone partitioning strategies can be selected based on the map's characteristics, while overall trends in performance remain consistent.
Finally, path smoothing is the last step of the algorithm. As shown in Table IV, shortcut-based methods achieve the highest improvement in path length, reducing it by approximately 30\%, as also illustrated in Figure 5.

Additionally, The Zonal RL-RRT algorithm is evaluated against a variety of recent heuristic and learning-based approaches, including the heuristic RRT*J \cite{b59}, the learning-based algorithms MPNet \cite{b24} and Neural RRT \cite{b26}. The evaluation focuses on running time efficiency and cost as reported in their respective studies. In each study, 4 number of different maps were used to assess the algorithms' running time performance. As shown in Table II, we compared the performance our algorithm to these approaches across all proposed maps. Figure 2 provides visual representations of the generated maps. Our results demonstrate that Zonal RL-RRT achieves competitive running time efficiency and cost compared to both heuristic and learning-based algorithms. 

\begin{figure}[ht] 
\centering

\begin{tabular}{c c c}
     \hspace{1.5em} \textbf{Map 1} \hspace{3.5em} & \textbf{Map 2} \hspace{3.5em} & \textbf{Map 3} \\
\end{tabular}

\vspace{1em} 

\begin{tabular}{c c c} 
    \textit{RRT*J} & \textit{NRRT*-S2} & \textit{MPNetSMP} \\
    \fbox{\includegraphics[width=0.12\textwidth, trim={1 4 5 5}, clip]{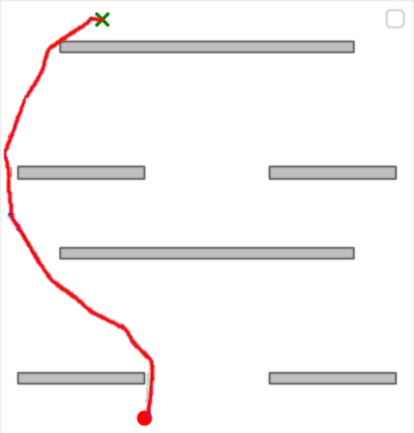}} &
    \fbox{\includegraphics[width=0.12\textwidth,trim={1 4 5 5}, clip]{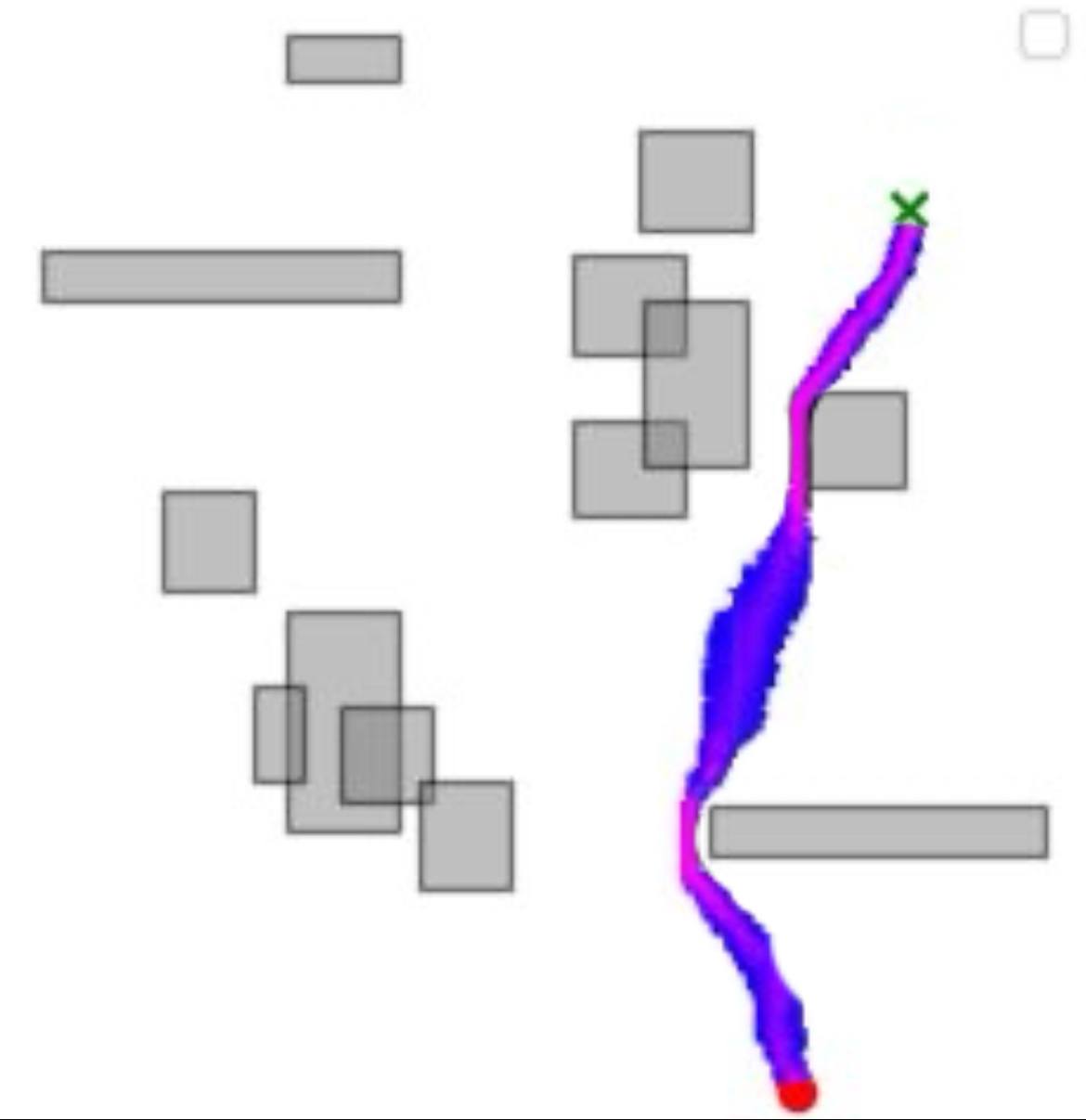}} &
    \fbox{\includegraphics[width=0.12\textwidth ,trim={1 4 2 1}, clip]{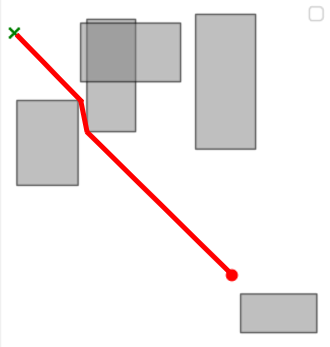}} 
    \vspace{0.1cm} \\
    (a) t = 0.058s & (b) t = 0.095s & (c) t = 0.210s \\
    \vspace{-0.3cm} \\  
    \textit{ZRL-RRT} & \textit{ZRL-RRT} & \textit{ZRL-RRT} \\
    \fbox{\includegraphics[width=0.12\textwidth,trim={1 4 1 5}, clip]{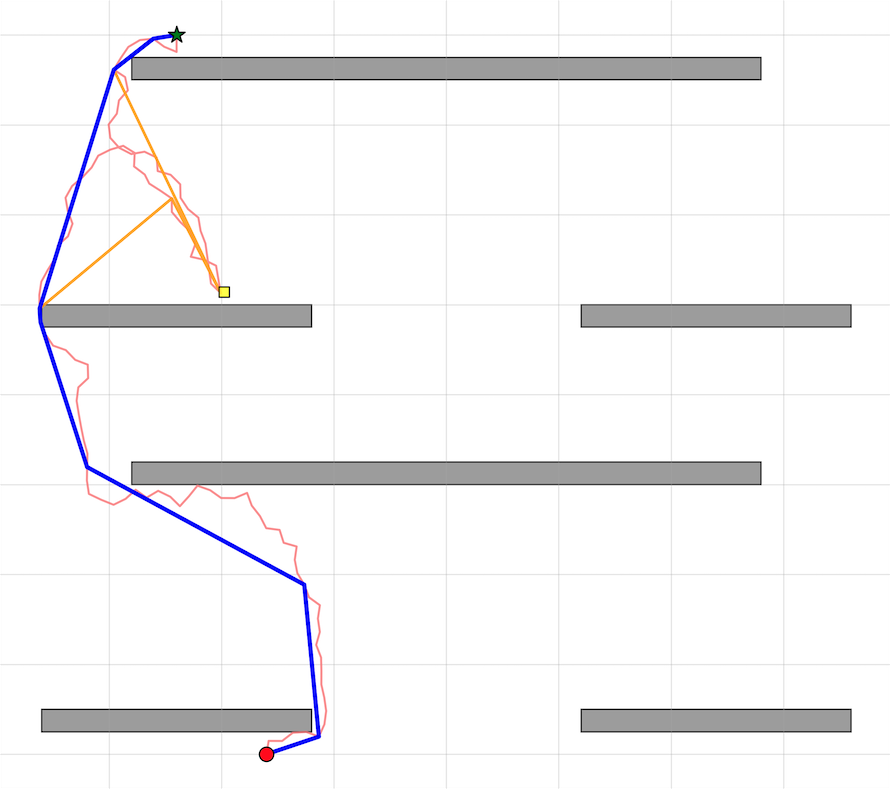}} &
    \fbox{\includegraphics[width=0.12\textwidth,trim={1 4 1 5}, clip]{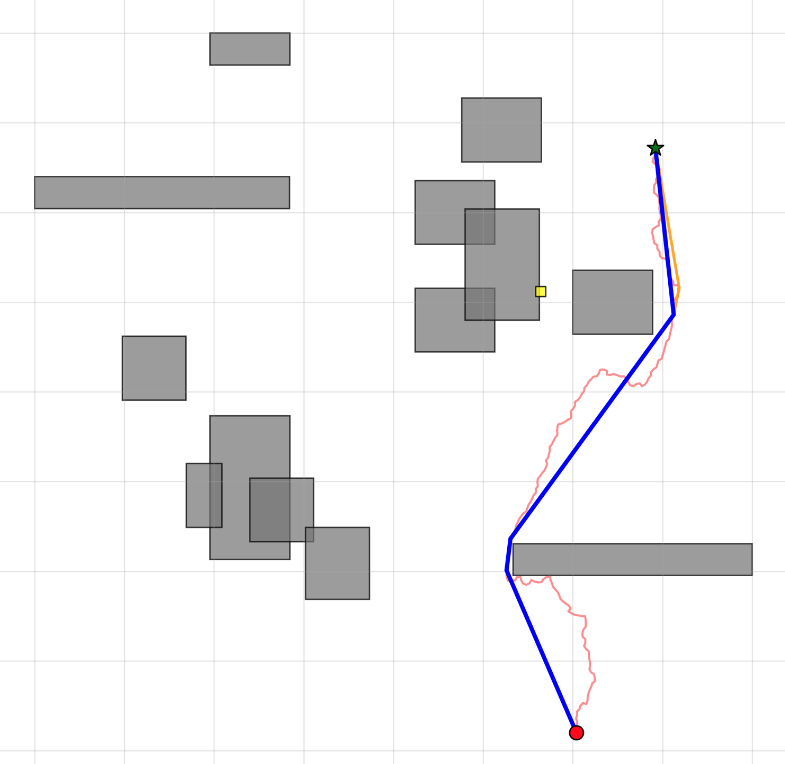}} &
    \fbox{\includegraphics[width=0.12\textwidth,trim={1 4 1 5}, clip]{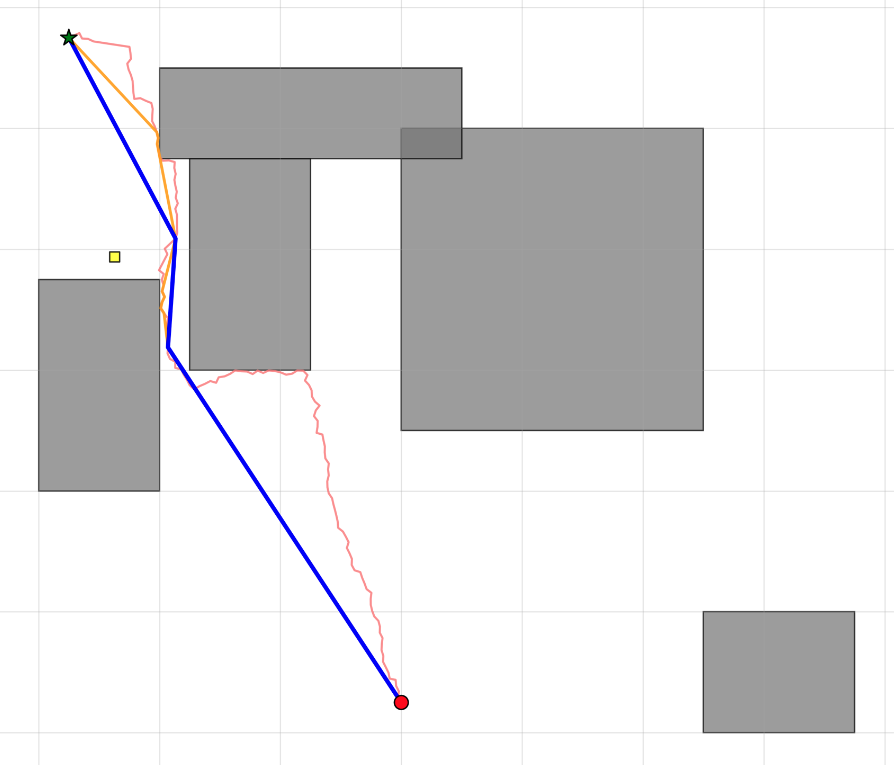}} 
    \vspace{0.0cm} \\
    (d) t = 0.052s & (e) t = 0.074s & (f) t = 0.019s \\
\end{tabular}

\caption{\small \textit{Comparative analysis of path planning algorithms in three different map scenarios. Top row: performance of baseline planners, including RRT*J (a), NRRT*-S2 (b), and MPNetSMP (c). Bottom row: performance of the proposed Zonal RL-RRT algorithm for the same maps (d, e, f). The red trajectory corresponds to the baseline RRT solution. The orange trajectory shows the result of applying smoothing independently within each zone. The blue trajectory represents the final path after global shortcut smoothing across the entire trajectory, producing a more coherent and shorter path.}}
\label{fig:example2}
\vspace{-0.4cm}
\end{figure}
\begin{table}
\centering
\caption{Comparison of Algorithms Based on Average Time}
\label{my-label}
\small
\begin{tabular}{@{}llll@{}}
\toprule
Dimension & Algorithm & Time(\%) & cost(\%)\\ \midrule
2D & MPNet SMP   &     \\ 
& \textbf{ZRL-RRT (Ours)}    & +50.0   &  -0.4\%\\ \midrule
2D & NRRT*-S6   &    \\ 
& \textbf{ZRL-RRT,  (Ours)}    & +43.9 &  -3.0\%    \\ \midrule
2D & RRT*J   &    \\ 
& \textbf{ZRL-RRT}    & +50.2  & -8.4\%   \\ \midrule
3D & MPNet SMP   &    \\ 
& \textbf{ZRL-RRT}    & +43.5   & +1.4\%   \\ \midrule
6D & BIT*  & 3.30    \\ 
& RRT*  & 5.55      \\
& \textbf{ZRL-RRT, depth=4 (Ours)}    & 3.09     \\  \midrule
\end{tabular}
\vspace{-0.3cm}
\end{table}
\begin{figure}[ht] 
\centering

\begin{tabular}{c c}
    \hspace{0em} \textbf{Map 1} \hspace{5em} & \textbf{Map 2} \\
\end{tabular}

\vspace{0.5em} 

\begin{tabular}{c c} 
    \textit{MPNetSMP}  & \textit{Zonal RL-RRT} \\
    \fbox{\includegraphics[width=0.15\textwidth]{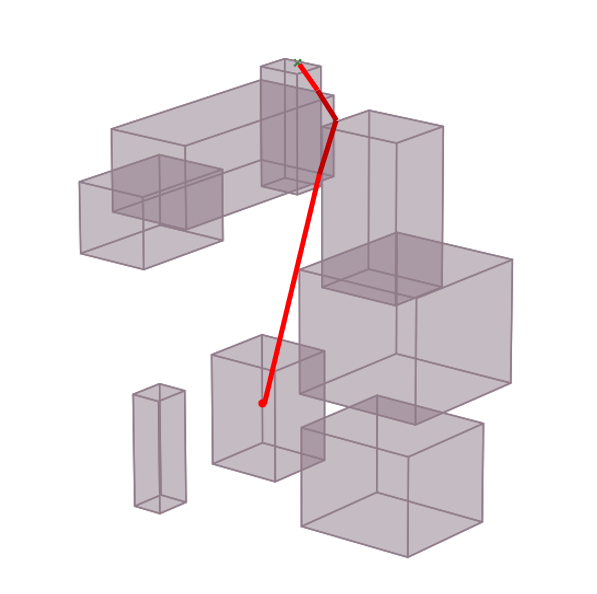}} &
    \fbox{\includegraphics[width=0.15\textwidth]{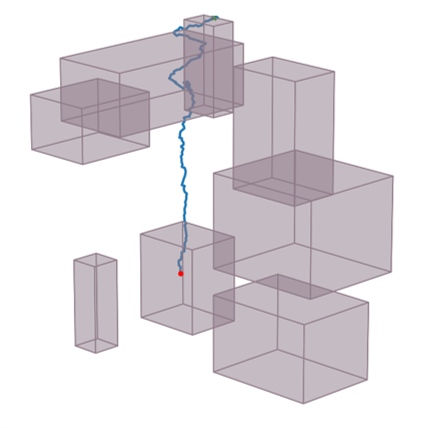}} \\
    \\
\end{tabular}
\caption{\small \textit{Comparative analysis of path planning algorithms on two 3D maps.}}
\label{fig:example3}
\vspace{-0.6cm}
\end{figure}

\subsection{3D Environment}
\vspace{-.05cm}
In this evaluation, we assessed the performance of Zonal RL-RRT against baseline algorithms in a \(200 \times 200 \times 200\) 3D map populated with randomly distributed box-shaped obstacles (see Table IV). To extend the kd-tree partitioning into 3D, we adjusted the depth and splitting strategy accordingly. For both depths 3 and 4, the Z-axis was further divided using the medians of the Z coordinates, resulting in 32 and 64 zones, respectively. The algorithms exhibited behavior similar to that observed in the 2D random maps from the previous section. On average, Zonal RL-RRT with depth 4 is about 1.9× faster than RRT and 2.5× faster than BIT*, and it also outpaces both RRT* and Informed RRT* in runtime. In terms of path length, Zonal RL-RRT achieves a 14.3\% improvement compared to RRT and RRT*. Compared to informed sampling–based planners, its cost remains within 2\% of BIT* and within 12.3\% of Informed RRT*, while also maintaining higher success rates.
To compare with the MPNet algorithm, the results are presented in Table II, and an example of the generated paths is shown in Figure 3. In 3D environments with depth 4, which results in 64 zones, the high-level planning components, such as zone connectivity and value iteration, incur at least a 4 times increase in time (on the order of 10 ms); however, their combined share remains around 1\% of the total planning time, with the low-level planner continuing to dominate the overall computation cost.

\subsection{Flexibility in Strategy Selection}
\vspace{-.05cm}
The Learning-based framework for high-level planning over the zones lays the groundwork for incorporating different strategies. Figure 5 clearly illustrates the different types of strategies we can employ. By setting \(R_d\) to zero and keeping the \(R_{\rho}\) parameter non-zero, thereby prioritizing the avoidance of zones with higher collision probability, the agent navigates around obstacles and prefers traversing through free zones. Conversely, by setting the \(R_{\rho}\) parameter to zero and keeping \(R_d\) non-zero, the agent adopts a more direct, goal-oriented approach, moving through zones closer to the goal. For all the scenarios in Table 1, setting the \( R_p \) parameter to zero improves the distance metric by up to 30\% and shows slightly better performance than RRT*. This flexibility and adaptability in strategy, based on diverse environmental configurations, highlight the potential of our algorithm to efficiently handle a wide range of scenarios and requirements. 

\begin{figure}[h]
\centering
\begin{tikzpicture}
\begin{axis}[
    ybar stacked,
    width=8cm,
    height=6cm,
    xlabel={Depth},
    ylabel={Time (seconds)},
    title={Algorithm Performance Breakdown by Depth},
    symbolic x coords={Depth 2, Depth 3, Depth 4},
    xtick=data,
    ymin=0,
    ymax=0.6,
    legend style={at={(0.98,0.98)}, anchor=north east, font=\tiny},
    bar width=1.5cm,
    area legend,
    x tick label style={font=\small},
    y tick label style={font=\small},
]

\addplot[fill=blue!60, draw=blue!80, line width=0.5pt] coordinates {
    (Depth 2, 0.001)   
    (Depth 3, 0.003)   
    (Depth 4, 0.016)   
};

\addplot[fill=green!60, draw=green!80, line width=0.5pt] coordinates {
    (Depth 2, 0.00556)   
    (Depth 3, 0.011)   
    (Depth 4, 0.020)   
};

\addplot[fill=red!50, draw=red!70, line width=0.5pt] coordinates {
    (Depth 2, 0.407324)  
    (Depth 3, 0.157504)  
    (Depth 4, 0.105821)  
};

\addplot[fill=orange!50, draw=orange!70, line width=0.5pt] coordinates {
    (Depth 2, 0.13756)   
    (Depth 3, 0.081044)  
    (Depth 4, 0.068411)  
};

\legend{Zone Connectivity, Value Iteration, RRT, Smoothing}

\end{axis}
\end{tikzpicture}
\caption{Breakdown of algorithm performance across different depths showing both absolute timing and relative proportions.}
\label{fig:algorithm_breakdown}
\end{figure}

For high-level planning, different algorithms can be implemented depending on the problem and its constraints. Our experiments over the scenarios reported in our study show that, on average, the A* algorithm performs about 10 times faster than Value Iteration due to its heuristic nature, solving only the shortest path from the start zone to the goal zone. However, the additional computational overhead of the high-level planner is minor (less than 1\%) compared to the low level planning process (Algorithm 1), and our reported performance metrics remain largely unaffected. This observation demonstrates that employing a hierarchical planning structure provides robust high-level guidance without incurring significant overhead, thereby benefiting the overall performance. We opted for a learning-based approach over a purely heuristic one like A* because it offers greater potential for future studies requiring adaptive, multi-agent, and map-wide planning strategies.

\begin{figure}[h]
\centering
\begin{tikzpicture}
\begin{axis}[
    ybar stacked,
    width=8cm,
    height=6cm,
    xlabel={Depth},
    ylabel={Path Length (units)},
    title={Path Length: Original vs Smoothed},
    symbolic x coords={Depth 2, Depth 3, Depth 4},
    xtick=data,
    ymin=0,
    ymax=500,
    legend style={at={(0.98,0.98)}, anchor=north east, font=\tiny, 
                  legend cell align=left},
    bar width=1.2cm,
    area legend,
    x tick label style={font=\small},
    y tick label style={font=\small},
]

\addplot[fill=blue!50, draw=blue!70, line width=0.5pt] coordinates {
    (Depth 2, 334)  
    (Depth 3, 277)  
    (Depth 4, 267)  
};

\addplot[fill=red!50, draw=red!70, line width=0.5pt] coordinates {
    (Depth 2, 117)  
    (Depth 3, 97)   
    (Depth 4, 93)   
};

\legend{Smoothed, Original Addition}

\end{axis}
\end{tikzpicture}
\caption{Path length comparison showing smoothed paths and additional length in original paths}
\label{fig:path_length_comparison}
\end{figure}

\begin{table}
\centering
\caption{Evaluation of Smoothing algorithms: Performance, Runtime, and Consistency in 2D maps}
\label{my-label}
\small
\begin{tabular}{@{}lcc@{}}
\toprule
Algorithm & cost (\%) & Time (\%) \\ \midrule
greedy Shortcut & 30.6 & 16.5 \\
Douglas-Peucker & 18.0 & 30.3 \\
Bezier Curve & 8.8 & 30.0 \\
Gussian Filter & 1.7 & 1.3 \\ \midrule
16 zones & Sr (\%) & cost (\%) \\ \midrule
\textbf{kd-tree} & \textbf{100} & \\
UG & 100 & +10 \\
QuadTree  & 100 & +19
\end{tabular}
\vspace{-0.7cm}
\end{table}

\begin{figure}[ht]
\centering
\begin{tabular}{@{\hskip 12pt} c @{\hskip 12pt} c @{\hskip 12pt} c} 
    \includegraphics[scale=0.09]{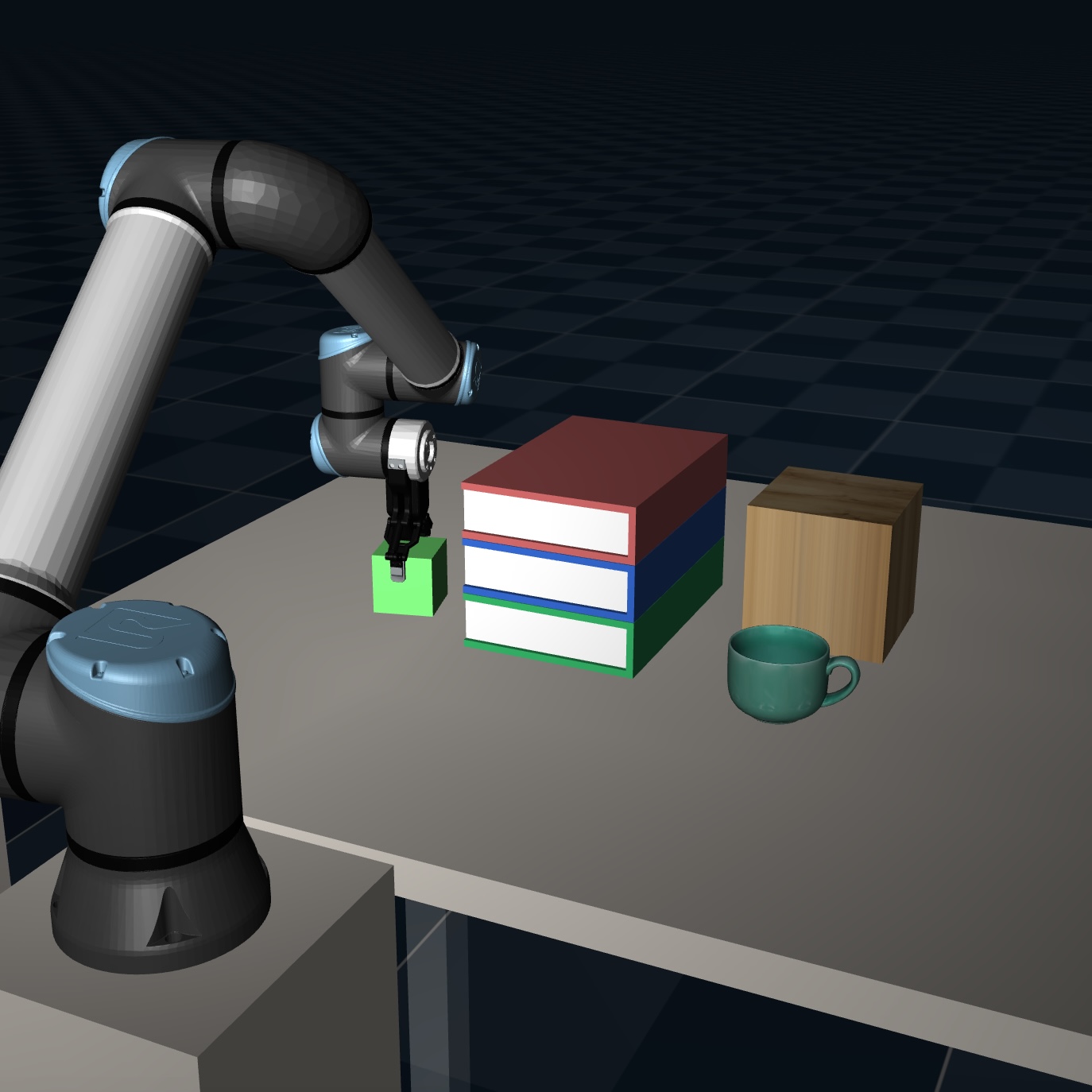} & 
    \includegraphics[scale=0.092]{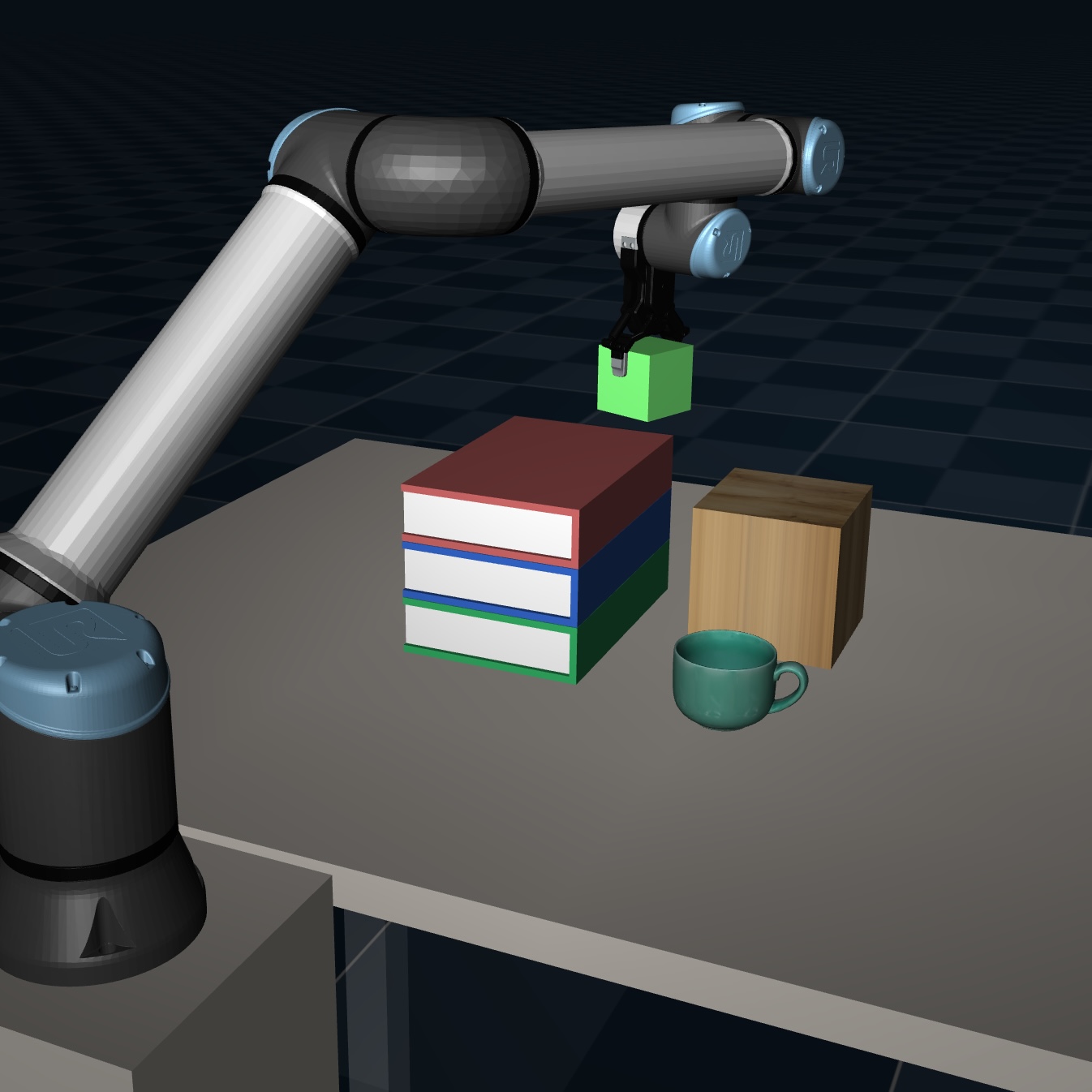} & 
    \includegraphics[scale=0.095]{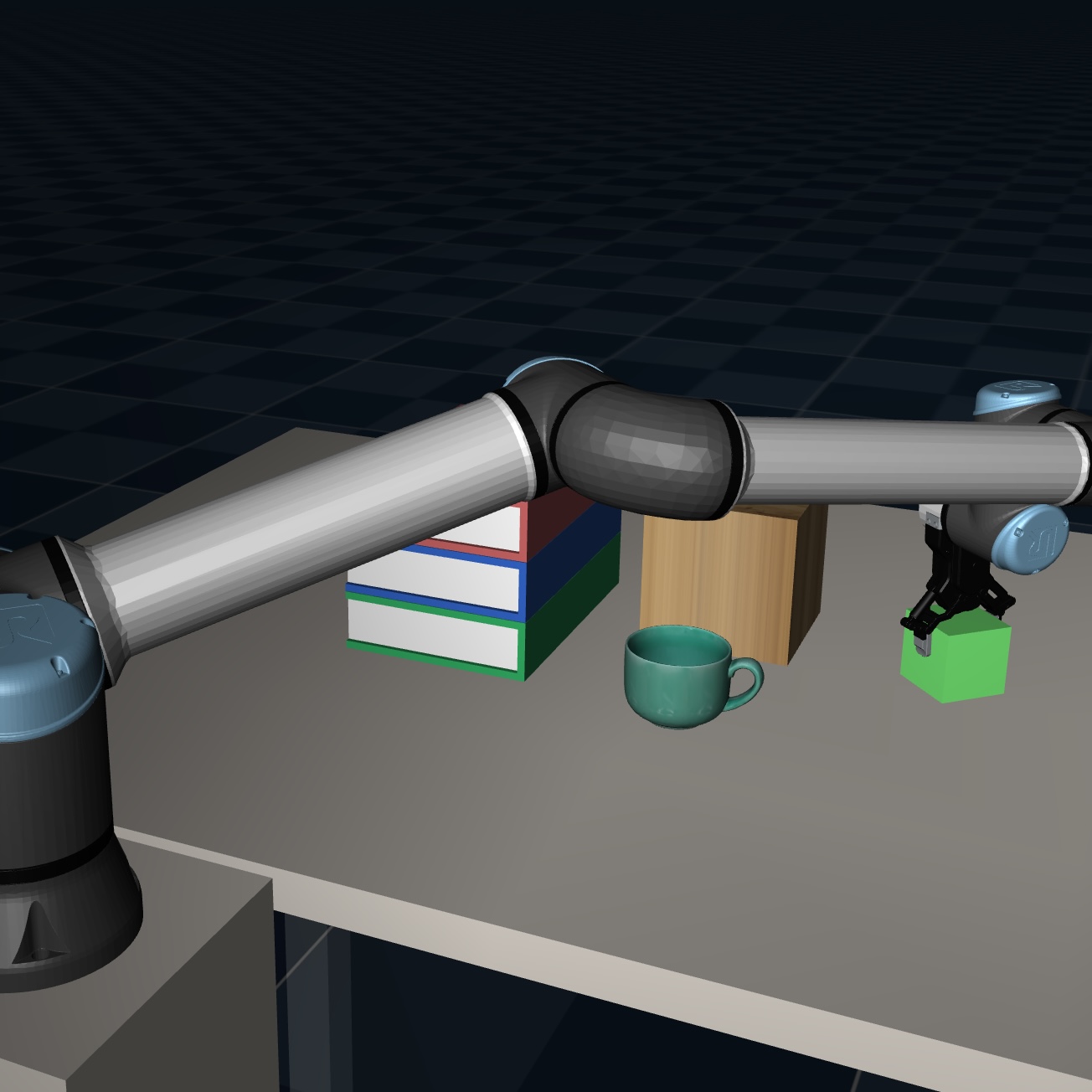} \\
    \includegraphics[scale=0.09]{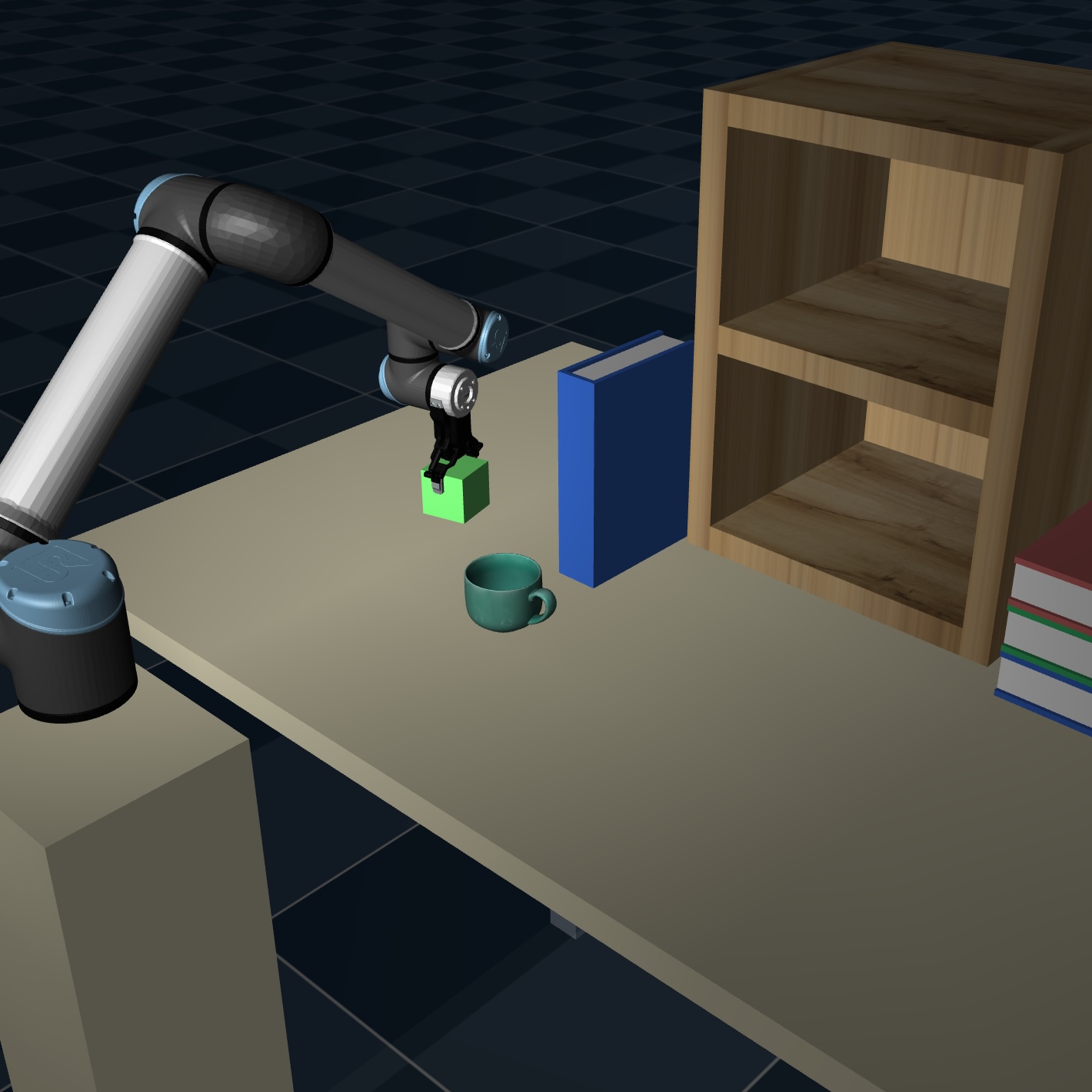} & 
    \includegraphics[scale=0.0905]{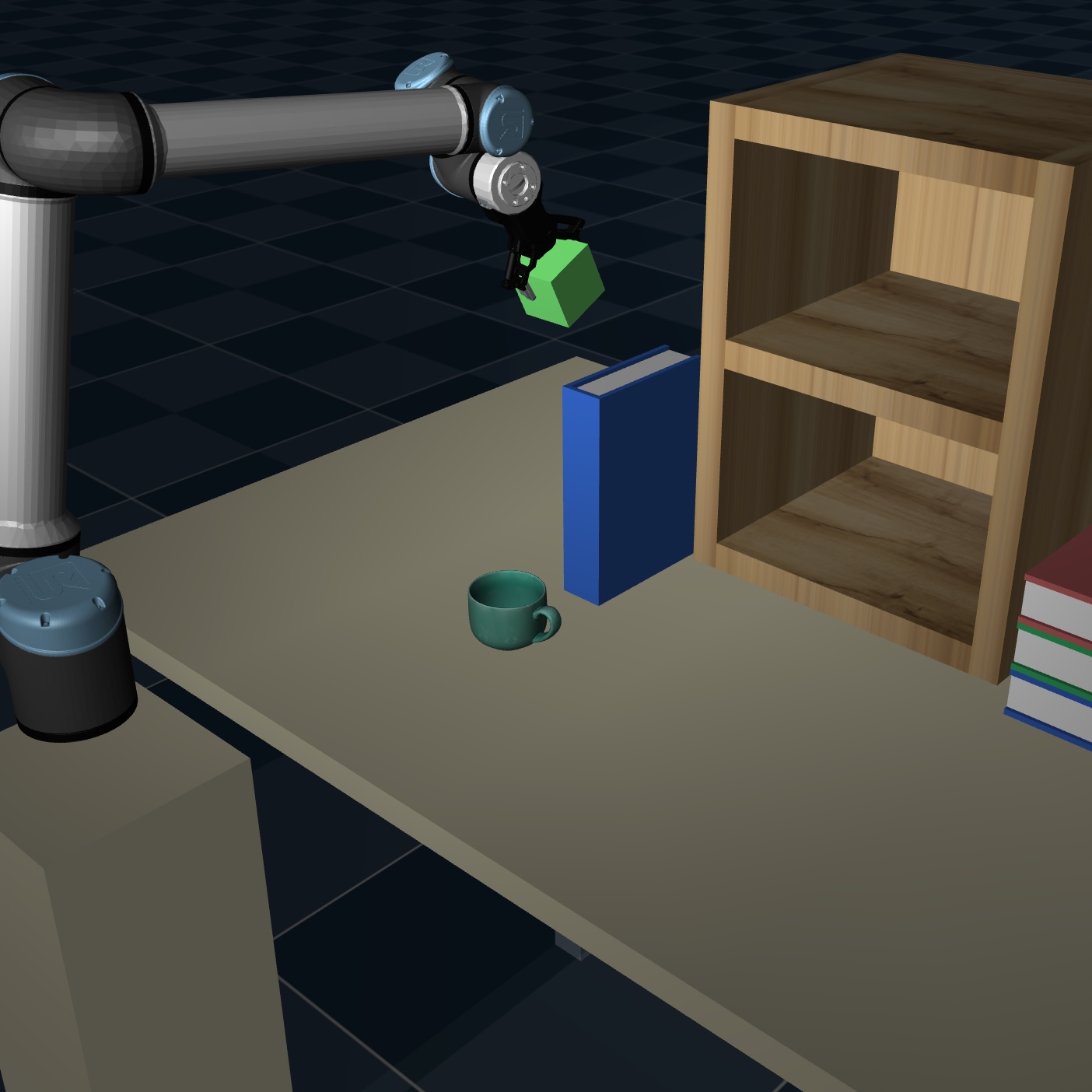} & 
    \includegraphics[scale=0.089]{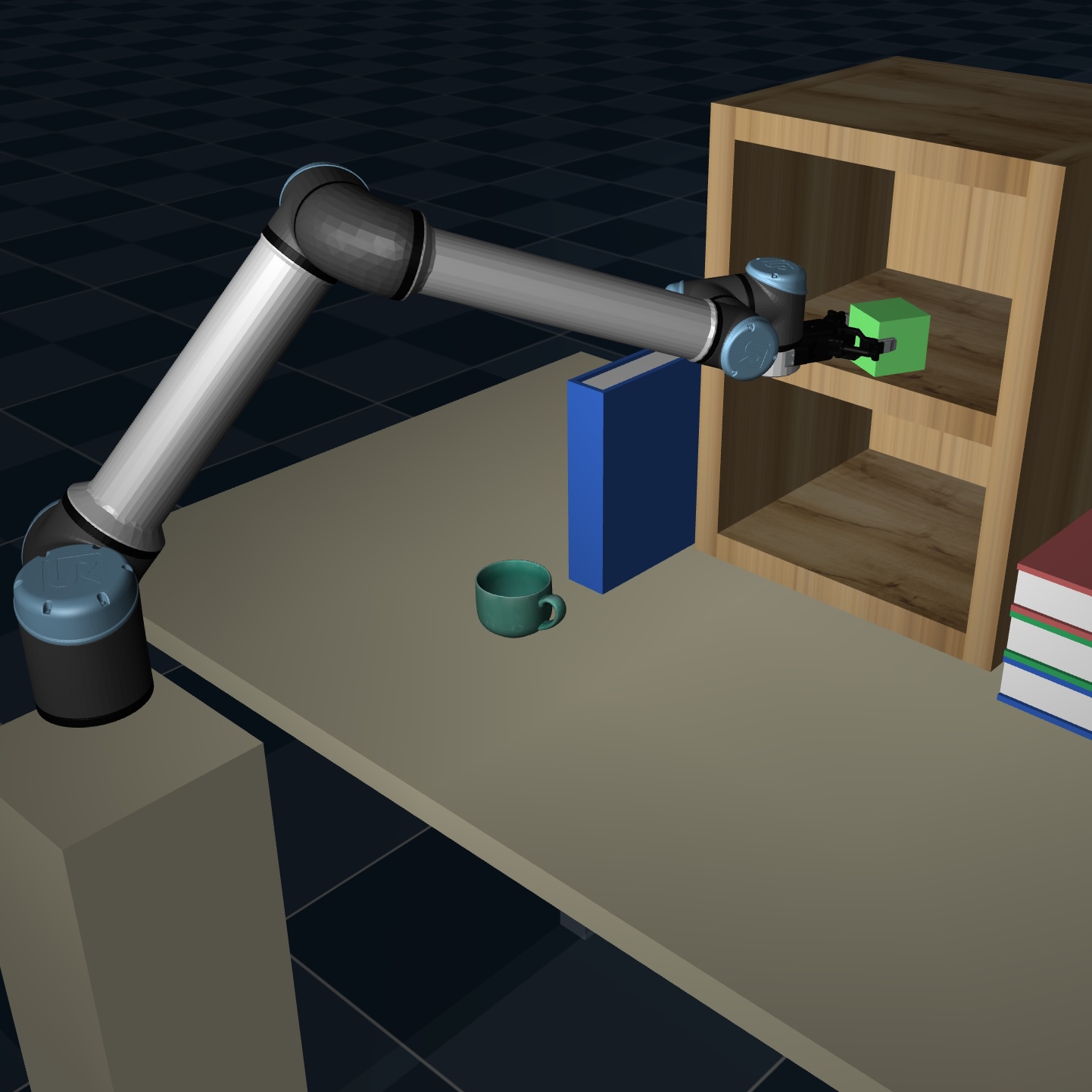} \\
\end{tabular} 
\caption{\small \textit{Two examples of the sequence of configurations during the path planning process for a UR10e robot arm in the MuJoCo environment.}}
\label{fig:example}
\vspace{-0.35cm}
\end{figure}

\subsection{Zonal RL-RRT on a 6-DOF Manipulator}
\vspace{-.05cm}
For further testing, we deployed Zonal RL-RRT on a UR10e robot arm with six degrees of freedom (DOF)~\cite{b64}, each corresponding to a rotating joint. The algorithms generate the trajectory for the robot's end effector in a 3D environment containing obstacles. Using the inverse kinematic formulation of the robot~\cite{b65} and MuJoCo’s contact detection capabilities~\cite{b66}, the validity of each state was verified. The performance of the algorithm was evaluated in five different environments, with the results presented in Table~II. The running times of BIT* are based on the initial path solutions, and Figure~4 illustrates two example simulations in MuJoCo 3.2.3. In terms of the distance metric, the algorithm shows only a 6\% difference compared to BIT*, while still achieving a 20\% improvement over RRT* on average. 

    
    
    
    
    




\begin{table}[h]
\centering
\caption{Performance Comparison of Path Planning Algorithms in 3D Forest Map}
\label{my-label}
\small
\setlength{\tabcolsep}{3pt} 
\renewcommand{\arraystretch}{0.9} 
\begin{tabular}{@{}cclccl@{}} 
\toprule
No. Obs & Obs Size & \multicolumn{1}{c}{Algorithm} & Time(s) & Sr (\%)& Cost(\%) \\ \midrule
250    & 20   & \raggedleft RRT                        & 0.67  & 100 & +19 \\  
       &      & \raggedleft RRT*                       &  30.05 & 100 & +18 \\  
       &      & \raggedleft BIT*                        & 0.69  & 100 & +1\\  
       &      & \raggedleft Informed RRT*              & 12.07  &  100 & -13\\  
       &      & \raggedleft \textbf{ZRL-RRT, depth=4} & \textbf{0.69}  & \textbf{100} &  \\  
       \midrule
1000     & 10   & \raggedleft RRT                        & 4.29  & 96  & +13\\  
       &      & \raggedleft RRT*                       & 14.07  & 100 & +12  \\  
       &      & \raggedleft BIT*                        & 6.04  & 100 & -3 \\  
       &      & \raggedleft Informed RRT*              & 45.10  & 97 & -15  \\  
       &      & \raggedleft \textbf{ZRL-RRT, depth=4 } & \textbf{2.40}  & \textbf{100} &   \\  
     \midrule
2000     & 5   & \raggedleft RRT                        & 7.79  & 84 & +14  \\  
       &      & \raggedleft RRT*                       & 30.07  &100  & +10 \\  
       &      & \raggedleft BIT*                        & 10.09  & 100 & -3 \\  
       &      & \raggedleft Informed RRT*              &  40.11 & 92 & -9 \\  
       &      & \raggedleft \textbf{ZRL-RRT, depth=4 } & \textbf{3.78}  & \textbf{100} &   \\  
       \midrule
\end{tabular}

\end{table}

\begin{figure}[hb] 
\centering

\begin{tabular}{c c}
    \textbf{Map 1} \hspace{6em} & \textbf{Map 2} \\
\end{tabular}

\vspace{0.2em} 

\begin{tabular}{c c} 
    \textit{ZRL-RRT: $R_{\rho}=0$} & \textit{ZRL-RRT: $R_{\rho}=0$} \\
    \fbox{\includegraphics[width=0.13\textwidth]{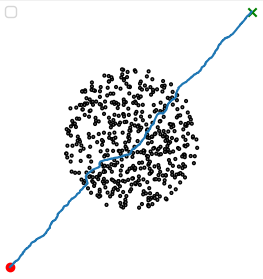}} &
    \fbox{\includegraphics[width=0.14\textwidth]{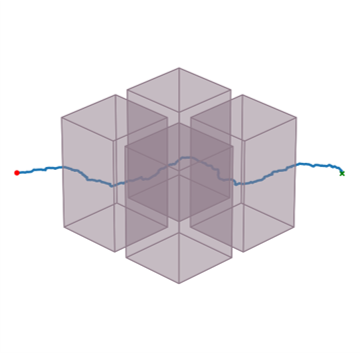}} \\
    \small (a) path length= 275 & (b) path length= 256 \\
    \textit{ZRL-RRT: $R_{d}=0$} & \textit{ZRL-RRT: $R_{d}=0$} \\
    \fbox{\includegraphics[width=0.14\textwidth]{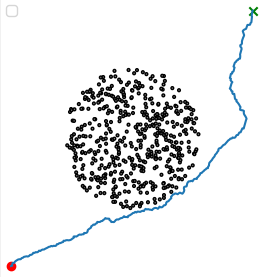}} &
    \fbox{\includegraphics[width=0.15\textwidth]{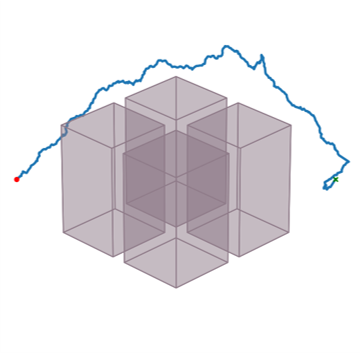}} \\
     (d) path length= 293 & (e) path length= 259 \\
\end{tabular}
\caption{\small \textit{Illustration of Zonal RL-RRT's flexibility with different rewards. In both maps, when $R_d \neq 0$, the agent takes a direct path, while $R_{\rho} \neq 0$ leads the agent around obstacles.}}
\label{fig:example3}
\vspace{-0.6cm}
\end{figure}
\section{Conclusion and future works}
\vspace{-.05cm}
In this work, we introduced Zonal RL-RRT, a novel path-planning algorithm designed for efficiently navigating diverse environments. Evaluations from 2D to 6D demonstrated that the algorithm consistently generates feasible paths with high success rates and competitive running times compared to baseline approaches. Future work could explore expanding the approach to multi-agent systems could unlock the algorithm’s high-level planning potential, enabling effective collaboration and coordination among agents in shared environments, which is crucial for cooperative tasks in scenarios such as large-scale autonomous vehicle fleets.


\begin{thebibliography}{00}

\bibitem{b1} Kim, Sanghyun and Park, Jongmin and Yun, Jae-Kwan and Seo, Jiwon. (2020). Motion Planning by Reinforcement Learning for an Unmanned Aerial Vehicle in Virtual Open Space with Static Obstacles. 
\bibitem{b2} Yan, Chao \& Xiaojia, Xiang \& Wang, Chang. (2020). Towards Real-Time Path Planning through Deep Reinforcement Learning for a UAV in Dynamic Environments. Journal of Intelligent \& Robotic Systems. 98. 10.1007/s10846-019-01073-3. 
\bibitem{b3} Yang, Laiyi \& Bi, Jing \& Yuan, Haitao. (2022). Dynamic Path Planning for Mobile Robots with Deep Reinforcement Learning. IFAC-PapersOnLine. 55. 19-24. 10.1016/j.ifacol.2022.08.042. 
\bibitem{b4} Bae, Hyansu, Gidong Kim, Jonguk Kim, Dianwei Qian, and Sukgyu Lee. 2019. "Multi-Robot Path Planning Method Using Reinforcement Learning" Applied Sciences 9, no. 15: 3057. https://doi.org/10.3390/app9153057
\bibitem{b7} A. Yahja, A. Stentz, S. Singh and B. L. Brumitt, "Framed-quadtree path planning for mobile robots operating in sparse environments," Proceedings. 1998 IEEE International Conference on Robotics and Automation (Cat. No.98CH36146), Leuven, Belgium, 1998, pp. 650-655 vol.1.
\bibitem{b8} Debnath, Sanjoy \& Omar, Rosli \& Bagchi, Susama \& elia nadira, Sabudin \& Shee Kandar, Mohd Haris Asyraf \& Foysol, K. \& Chakraborty, Tapan. (2021). Different Cell Decomposition Path Planning Methods for Unmanned Air Vehicles-A Review. 10.1007/978-981-15-5281-6\_8. 
\bibitem{b9} Wang, Chien-Yen \& Yang, Chan-Yun \& Banitaan, Shadi \& Luo, Chaomin \& Galsanbadam, Sainzaya. (2019). Coarse grid partition to speed up A* robot navigation. Journal of the Chinese Institute of Engineers. 43. 1-14. 10.1080/02533839.2019.1694444. 
\bibitem{b10} LaValle, Steven M.. “Rapidly-exploring random trees : a new tool for path planning.” The annual research report (1998): n. pag.
\bibitem{b11} J. J. Kuffner and S. M. LaValle, "RRT-connect: An efficient approach to single-query path planning," Proceedings 2000 ICRA. Millennium Conference. IEEE International Conference on Robotics and Automation. Symposia Proceedings (Cat. No.00CH37065), San Francisco, CA, USA, 2000, pp. 995-1001 vol.2.
\bibitem{b12} M. Elbanhawi and M. Simic, "Sampling-Based Robot Motion Planning: A Review," in IEEE Access, vol. 2, pp. 56-77, 2014, doi: 10.1109/ACCESS.2014.2302442.
\bibitem{b13} J. D. Gammell, S. S. Srinivasa and T. D. Barfoot, "Batch Informed Trees (BIT*): Sampling-based optimal planning via the heuristically guided search of implicit random geometric graphs," 2015 IEEE International Conference on Robotics and Automation (ICRA).
\bibitem{b14} L. Palmieri, S. Koenig and K. O. Arras, "RRT-based nonholonomic motion planning using any-angle path biasing," 2016 IEEE International Conference on Robotics and Automation (ICRA), Stockholm, Sweden, 2016.
\bibitem{b15} J. D. Gammell, S. S. Srinivasa and T. D. Barfoot, "Batch Informed Trees (BIT*): Sampling-based optimal planning via the heuristically guided search of implicit random geometric graphs," 2015 IEEE International Conference on Robotics and Automation (ICRA), Seattle, WA, USA, 2015.
\bibitem{b16} Brunner, Michael \& Brüggemann, Bernd \& Schulz, Dirk. (2013). Hierarchical Rough Terrain Motion Planning using an Optimal Sampling-Based Method. Proceedings - IEEE International Conference on Robotics and Automation. 10.1109/ICRA.2013.6631372. 
\bibitem{b17} H. An, J. Hu and P. Lou, "Obstacle Avoidance Path Planning Based on Improved APF and RRT," 2021 4th International Conference on Advanced Electronic Materials, Computers and Software Engineering (AEMCSE), Changsha, China, 2021, pp. 1028-1032, doi: 10.1109/AEMCSE51986.2021.00210.
\bibitem{b18} Qureshi, A.H., Ayaz, Y. Potential functions based sampling heuristic for optimal path planning. Auton Robot 40, 1079–1093 (2016). https://doi.org/10.1007/s10514-015-9518-0
\bibitem{b19} Kober, Jens \& Bagnell, J. \& Peters, Jan. (2013). Reinforcement Learning in Robotics: A Survey. The International Journal of Robotics Research. 32. 1238-1274. 10.1177/0278364913495721. 
\bibitem{b20} L. Dong, Z. He, C. Song and C. Sun, "A review of mobile robot motion planning methods: from classical motion planning workflows to reinforcement learning-based architectures," in Journal of Systems Engineering and Electronics, vol. 34.
\bibitem{b21}Kaelbling, Leslie Pack, Michael L. Littman and Andrew W. Moore. “Reinforcement Learning: A Survey.” J. Artif. Intell. Res. 4 (1996): 237-285.
\bibitem{b22}M. Zucker, J. Kuffner and J. A. Bagnell, "Adaptive workspace biasing for sampling-based planners," 2008 IEEE International Conference on Robotics and Automation, Pasadena, CA, USA, 2008, pp. 3757-3762, doi: 10.1109/ROBOT.2008.4543787.
\bibitem{b23}Brian Ichter, James Harrison, and Marco Pavone. 2018. Learning Sampling Distributions for Robot Motion Planning. In 2018 IEEE International Conference on Robotics and Automation (ICRA). IEEE Press, 7087–7094. 
\bibitem{b24} Qureshi, A. H., Yinglong Miao, Anthony Simeonov and Michael C. Yip. “Motion Planning Networks: Bridging the Gap Between Learning-Based and Classical Motion Planners.” IEEE Transactions on Robotics 37 (2019): 48-66.
\bibitem{b26}J. Wang, W. Chi, C. Li, C. Wang and M. Q. . -H. Meng, "Neural RRT*: Learning-Based Optimal Path Planning," in IEEE Transactions on Automation Science and Engineering, vol. 17, no. 4, pp. 1748-1758, Oct. 2020, doi: 10.1109/TASE.2020.2976560.
\bibitem{b27}Y. Li, R. Cui, Z. Li and D. Xu, "Neural Network Approximation Based Near-Optimal Motion Planning With Kinodynamic Constraints Using RRT," in IEEE Transactions on Industrial Electronics, vol. 65, no. 11, pp. 8718-8729, Nov. 2018, doi: 10.1109/TIE.2018.2816000.
\bibitem{b28}R. Franceschini, M. Fumagalli and J. C. Becerra, "Learn to efficiently exploit cost maps by combining RRT* with Reinforcement Learning," 2022 IEEE International Symposium on Safety, Security, and Rescue Robotics (SSRR), Sevilla, Spain, 2022, pp. 251-256, doi: 10.1109/SSRR56537.2022.10018735. 
\bibitem{b29} A. Konar, I. Goswami Chakraborty, S. J. Singh, L. C. Jain and A. K. Nagar, "A Deterministic Improved Q-Learning for Path Planning of a Mobile Robot," in IEEE Transactions on Systems, Man, and Cybernetics: Systems, vol. 43, no. 5, pp. 1141-1153, Sept. 2013.
\bibitem{b30} Ee Soong Low, Pauline Ong, Kah Chun Cheah, Solving the optimal path planning of a mobile robot using improved Q-learning, Robotics and Autonomous Systems, Volume 115, 2019.
\bibitem{b31} Abderraouf Maoudj, Abdelfetah Hentout, Optimal path planning approach based on Q-learning algorithm for mobile robots, Applied Soft Computing,2020.
\bibitem{b32} C. Yan and X. Xiang, "A Path Planning Algorithm for UAV Based on Improved Q-Learning," 2018 2nd International Conference on Robotics and Automation Sciences (ICRAS), Wuhan, China, 2018, pp. 1-5.
\bibitem{b33} S. Li, X. Xu and L. Zuo, "Dynamic path planning of a mobile robot with improved Q-learning algorithm," 2015 IEEE International Conference on Information and Automation, Lijiang, China, 2015, pp. 409-414, doi: 10.1109/ICInfA.2015.7279322.
\bibitem{b37}L. C. Garaffa, M. Basso, A. A. Konzen and E. P. de Freitas, "Reinforcement Learning for Mobile Robotics Exploration: A Survey," in IEEE Transactions on Neural Networks and Learning Systems, vol. 34, no. 8, pp. 3796-3810, Aug. 2023.
\bibitem{b38} Jens Kober, Andrew Bagnell and Jan Peters, Reinforcement Learning in Robotics: A survey, The International Journal of Robotics Research, Sep. 2013. 
\bibitem{b39}Z. Wang, Z. Hu, Y. Yang and Y. Yin, "Research on PPO algorithm in solving AUV path planning problems," 2021 2nd International Seminar on Artificial Intelligence, Networking and Information Technology (AINIT), Shanghai, China, 2021, pp. 73-79, doi: 10.1109/AINIT54228.2021.00024.
\bibitem{b40}Z. Wang, Z. Hu, Y. Yang and Y. Yin, "Research on PPO algorithm in solving AUV path planning problems," 2021 2nd International Seminar on Artificial Intelligence, Networking and Information Technology (AINIT), Shanghai, China, 2021, pp. 73-79, doi: 10.1109/AINIT54228.2021.00024. 
\bibitem{b41}Y. Dong and X. Zou, "Mobile Robot Path Planning Based on Improved DDPG Reinforcement Learning Algorithm," 2020 IEEE 11th International Conference on Software Engineering and Service Science (ICSESS), Beijing, China, 2020, pp. 52-56.
\bibitem{b42}Y. Zhao, X. Wang, R. Wang, Y. Yang and F. Lv, "Path Planning for Mobile Robots Based on TPR-DDPG," 2021 International Joint Conference on Neural Networks (IJCNN), Shenzhen, China, 2021, pp. 1-8, doi: 10.1109/IJCNN52387.2021.9533570. 
\bibitem{b43}A. Azzalini \& A. Capitanio, 1999. "Statistical applications of the multivariate skew normal distribution," Journal of the Royal Statistical Society Series B, Royal Statistical Society, vol. 61(3), pages 579-602.
\bibitem{b44}B. Siciliano and O. Khatib, Eds., Springer Handbook of Robotics. Springer, 2016.
\bibitem{b45}Huijuan Wang, Yuan Yu and Quanbo Yuan, "Application of Dijkstra algorithm in robot path-planning," 2011 Second International Conference on Mechanic Automation and Control Engineering, Hohhot, 2011.
\bibitem{b46}O. Souissi, R. Benatitallah, D. Duvivier, A. Artiba, N. Belanger and P. Feyzeau, "Path planning: A 2013 survey," Proceedings of 2013 International Conference on Industrial Engineering and Systems Management (IESM), Agdal, Morocco, 2013, pp. 1-8. 
\bibitem{b47}P. E. Hart, N. J. Nilsson and B. Raphael, "A Formal Basis for the Heuristic Determination of Minimum Cost Paths," in IEEE Transactions on Systems Science and Cybernetics, vol. 4, no. 2, pp. 100-107, July 1968, doi: 10.1109/TSSC.1968.300136.
\bibitem{b48}Khatib, Oussama. “Real-Time Obstacle Avoidance for Manipulators and Mobile Robots.” The International Journal of Robotics Research 5 (1985): 90 - 98.
\bibitem{b49}I. Baldwin and P. Newman, “Non-parametric learning for natural plan generation,” in Proc. IEEE/RSJ Int. Conf. Intell. Robots Syst., Oct. 2010, pp. 4311–4317.
\bibitem{b50}Brian Ichter, James Harrison, and Marco Pavone. 2018. Learning Sampling Distributions for Robot Motion Planning. In 2018 IEEE International Conference on Robotics and Automation (ICRA). IEEE Press, 7087–7094. https://doi.org/10.1109/ICRA.2018.8460730
\bibitem{b51}Bency, Mayur Joseph, Ahmed Hussain Qureshi and Michael C. Yip. “Neural Path Planning: Fixed Time, Near-Optimal Path Generation via Oracle Imitation.” 2019 IEEE/RSJ International Conference on Intelligent Robots and Systems (IROS) (2019): 3965-3972.
\bibitem{b52}Yonetani, Ryo, Tatsunori Taniai, Mohammadamin Barekatain, Mai Nishimura and Asako Kanezaki. “Path Planning using Neural A* Search.” International Conference on Machine Learning (2020).
\bibitem{b53}Bhardwaj, Mohak, Sanjiban Choudhury and Sebastian A. Scherer. “Learning Heuristic Search via Imitation.” Conference on Robot Learning (2017).
\bibitem{b54}Karaman, Sertac \& Frazzoli, Emilio. (2011). Sampling-based Algorithms for Optimal Motion Planning. International Journal of Robotic Research - IJRR. 30. 846-894. 10.1177/0278364911406761. 
\bibitem{b55}Luigi Palmieri, Sven Koenig, and Kai O. Arras. 2016. RRT-based nonholonomic motion planning using any-angle path biasing. In 2016 IEEE International Conference on Robotics and Automation (ICRA)
\bibitem{b56}Chiang, Hao-Tien Lewis, Jasmine Hsu, Marek Fiser, Lydia Tapia and Aleksandra Faust. “RL-RRT: Kinodynamic Motion Planning via Learning Reachability Estimators From RL Policies.” IEEE Robotics and Automation Letters 4 (2019): 4298-4305.
\bibitem{b57}J. D. Gammell, S. S. Srinivasa and T. D. Barfoot, "Informed RRT*: Optimal sampling-based path planning focused via direct sampling of an admissible ellipsoidal heuristic," 2014 IEEE/RSJ International Conference on Intelligent Robots and Systems, Chicago, IL, USA.
\bibitem{b58}Sakai, Atsushi, Daniel Ingram, Joseph Dinius, Karan Chawla, Antonin Raffin and Alexis Paques. “PythonRobotics: a Python code collection of robotics algorithms.” ArXiv abs/1808.10703 (2018): n. pag.
\bibitem{b59} I. A. Sucan, M. Moll and L. E. Kavraki, "The Open Motion Planning Library," in IEEE Robotics \& Automation Magazine, vol. 19, no. 4, pp. 72-82, Dec. 2012, doi: 10.1109/MRA.2012.2205651.

\bibitem{b60}J. Wang and M. Q. . -H. Meng, "Optimal Path Planning Using Generalized Voronoi Graph and Multiple Potential Functions," in IEEE Transactions on Industrial Electronics, vol. 67, no. 12, pp. 10621-10630, Dec. 2020.

\bibitem{b61} R. E. Bellman, "Dynamic Programming," Princeton University Press, 1957.

\bibitem{b62} D. P. Bertsekas, "Dynamic Programming and Optimal Control," vol. 1, 4th ed., Athena Scientific, 2012.

\bibitem{b63} R. S. Sutton and A. G. Barto, "Reinforcement Learning: An Introduction," 2nd ed., MIT Press, 2018.
\bibitem{b64} Universal Robots, “UR10e User Manual,” Universal Robots A/S, 2018. 
\bibitem{b65} "DH Parameters for Calculations of Kinematics and Dynamics", Universal Robots, [Online]. 
\bibitem{b67} E. Todorov, T. Erez, and Y. Tassa, “MuJoCo: A Physics Engine for Model-Based Control,” in \textit{2012 IEEE/RSJ International Conference on Intelligent Robots and Systems (IROS)}, pp. 5026–5033, 2012.
\bibitem{bruce2005}
J. Bruce, “Spatial Partitioning and kd‑Trees,” in \textit{Ph.D. Thesis, Carnegie Mellon University}, Pittsburgh, PA, 2005. Available: \url{https://www.cs.cmu.edu/~jbruce/thesis/chapters/thesis-ch03.pdf}
\bibitem{b65}
D. H. Douglas and T. K. Peucker, “Algorithms for the reduction of the number of points required to represent a digitized line or its caricature,” \textit{Cartographica: The International Journal for Geographic Information and Geovisualization}, vol. 10, no. 2, pp. 112–122, 1973.

\bibitem{b68}
Y. Liao et al., “Smooth path planning for mobile robots based on Bézier curve and improved A* algorithm,” in \textit{Proc. IEEE Int. Conf. Mechatronics and Automation (ICMA)}, pp. 2262–2267, 2018.






\end{thebibliography}
\end{document}